\documentclass[runningheads]{llncs}

\usepackage{eccv}

\usepackage{eccvabbrv}

\usepackage{graphicx}
\usepackage{booktabs}

\usepackage[accsupp]{axessibility}  

\usepackage{orcidlink}
\usepackage{tcolorbox}
\usepackage{algorithm}
\usepackage{algpseudocode}

\begin{document}

\title{BoardroomAI: Dependency-Aware Human-Steerable Multi-Agent Deliberation through Evolving Decision Graphs} 

\titlerunning{BoardroomAI: Human-Steerable Decision Graph Evolution}

\author{Sanjeev Manivannan\inst{1}}

\authorrunning{S.~Manivannan}

\institute{
Indian Institute of Technology Madras, Chennai, India\\
\email{sanjeev.manivannan@gmail.com}
}

\maketitle

\begin{abstract}
Organizational decisions are co-created while evidence, constraints, and human priorities continue to change. In conventional transcript-based multi-agent systems, a human typically provides an initial problem, agents deliberate internally, and the system returns a final response. BoardroomAI instead treats the human as a persistent participant who can intervene during the discussion by challenging assumptions, modifying constraints, changing priorities, introducing evidence, or redirecting the decision process. We operationalize this human--agent coexistence through four components: (i) a typed decision graph whose nodes represent evidence, assumptions, constraints, claims, objections, alternatives, risks, and decisions, and whose edges represent semantic dependencies and specialist responsibility; (ii) an intervention compiler that converts each confirmed human action into explicit node- and edge-level graph updates; (iii) a dependency-aware propagation mechanism that measures the structural impact of the intervention, identifies the affected decision subgraph, preserves unaffected artifacts, and selectively reactivates the relevant domain specialists; and (iv) an evaluation framework that measures intervention impact, repair coverage, preservation, recomputation, and final decision validity. On 600 generated decision-DAG interventions, the proposed propagation mechanism matched exhaustive impact computation while inspecting only 14.59\% of nodes. In a 12-case exploratory agent pilot, selective repair recomputed 62.11\% of canonical nodes, preserved all gold-unaffected nodes, and produced valid updated decisions in six cases, while conservatively abstaining in the remaining six. These abstentions show that correct intervention routing does not necessarily provide sufficient context for decision synthesis. We therefore formalize a decision-sufficient context closure for human-steered multi-agent deliberation. All reported results are synthetic and prototype-level.
\end{abstract}

\section{Introduction}
LLM agents can adopt roles, exchange critiques, and aggregate answers
\cite{du2024debate,li2023camel}. These capabilities do
not by themselves produce an accountable meeting. In a real decision process,
a chief executive may reduce a budget after several proposals have been
developed; legal evidence may be withdrawn; or an ethical objection may remain
unresolved despite a numerical majority. Restarting wastes work and can alter
unrelated conclusions through stochastic drift. Appending the change to a
transcript, however, offers no guarantee that every dependent artifact is
revisited.

The central research problem is \emph{revision routing}: after a human
intervention, the system must determine (i) which claims, assumptions,
constraints, or decisions have changed semantic status; (ii) which generated
artifacts must be revised, recomputed, or invalidated; (iii) which unaffected
artifacts should be preserved without unnecessary regeneration; and (iv) which
specialist agents must be reactivated to repair the affected portion of the
deliberation. This does not expose private
chain-of-thought. Instead, it maintains concise, externally generated
deliberation artifacts, including claims, evidence, assumptions, constraints,
objections, alternatives, dependencies, confidence estimates, and
provenance, that can be inspected, challenged, revised, and audited by the
human participant.

BoardroomAI treats the recommendation, its alternatives, and its supporting
rationale as a \emph{shared co-created artifact}. The human is neither a prompt
provider restricted to the beginning nor an approver restricted to the end and can redirect the process, revise constraints, challenge an inference,
change priorities, preserve a minority objection, or make an accountable
override while the agents continue to elaborate the artifact. Dependency-aware conversational revision and graph-based context selection
already exist \cite{he2026grounded,wang2026ledger}. BoardroomAI therefore
claims the following narrower system contribution:
\begin{enumerate}
    \item \textbf{Decision-graph representation.}
    We define a formal decision-graph semantics with alternative minimal
    justification environments for preserving independently supported claims.

    \item \textbf{Typed human interventions.}
    We formalize human interventions as typed graph operations, including
    invalidation, weakening, contestation, supersession, and preference changes.

    \item \textbf{Selective specialist repair.}
We propagate intervention impact through decision graph, selectively reactivate only affected specialists, and specify conservative fallback to full redeliberation when the repair is insufficient.

    \item \textbf{DynaBoard evaluation.}
    We introduce DynaBoard, a 12-case synthetic pilot for evaluating routing independent of the final decision quality.
\end{enumerate}
The current prototype is language-and-graph based rather than multimodal. Its
co-creation contribution lies in maintaining and revising an explicit shared
decision state as human intent evolves during deliberation. The present
evaluation examines (i) intervention-impact routing and preservation of
unaffected artifacts, (ii) decision-sufficient selective repair on complete and
corrupted dependency graphs, and (iii) controlled human interventions applied
through a programmatic controller. Unrestricted natural-language intervention
parsing, human comprehension, calibrated reliance, and robustness to ambiguous
or incorrect interventions remain subjects for larger agent evaluations and
human-subject studies.

\subsection{Motivation}

Let us look at a real-world boardroom situation to understand the motivation
behind our agentic framework.

\begin{tcolorbox}[
    colback=blue!10,
    colframe=blue!60,
    boxrule=0.5pt,
    arc=2mm,
    left=2mm,
    right=2mm,
    top=1mm,
    bottom=1mm]
\emph{Consider a company evaluating the launch of a new product. Finance,
marketing, legal, and operations specialists deliberate over multiple
alternatives and eventually recommend a staged product launch under a
100 million budget. After the discussion reaches consensus, the decision
maker reduces the budget to 40 million. While this change invalidates
some earlier conclusions, others remain fully justified. A
transcript-based multi-agent system either restarts the entire discussion
or continues without explicitly determining which previous decisions
remain valid, leading to unnecessary recomputation or inconsistent
reasoning.}
\end{tcolorbox}

BoardroomAI instead represents the deliberation as a dependency-aware decision
graph. The intervention is propagated only through affected dependencies,
invalidating the paid-acquisition launch strategy while preserving the
independent regional-partnership justification and unrelated legal
conclusions. Only the relevant specialists are reactivated, and the system
produces an auditable change log explaining what changed, why it changed, and
which decisions remain valid.

\section{Related Work}
\paragraph{Human--AI co-creation and mixed initiative.}
Co-creative systems study how humans and AI jointly contribute to an evolving
artifact through shared responsibility and interaction
\cite{kantosalo2016modes,rezwana2022cofi,lin2023ontology,lawton2023agency}.
BoardroomAI extends this perspective to organizational deliberation by treating
human interventions as explicit operations on a shared decision graph.

\paragraph{Multi-agent reasoning.}
Multi-agent debate and role-based collaboration improve reasoning, coordination,
and decision making across diverse tasks
\cite{du2024debate,li2023camel,wang2024rethinking,xiong2023interconsistency,turkstra2026argsbase,quan2026multicolleagues,ma2025deliberation,chiang2024devils,prakash2026dci}.
Unlike prior work, we study typed human interventions followed by
dependency-aware selective specialist repair.

\paragraph{Revision, argumentation, and provenance.}
Truth-maintenance systems, computational argumentation, and provenance models
provide formal mechanisms for representing dependencies, alternative
justifications, and explanation
\cite{dekleer1986atms,dung1995argumentation,conklin1988gibis,green2007provenance}.
BoardroomAI adapts these ideas to maintain a dynamically evolving
decision graph for human--AI deliberation.

\paragraph{Human control and group decision support.}
Mixed-initiative systems emphasize appropriate human oversight, while research
in decision support highlights the importance of calibrated trust, diverse
opinions, and preserving dissent
\cite{horvitz1999mixed,bansal2021whole,bucinca2021trust,stasser1985unshared,lorenz2011social}.
Accordingly, BoardroomAI treats dissent and human interventions as first-class,
auditable components of the deliberation process.

\section{Problem Formulation}
At turn \(t\), the deliberation state is
\[
\mathcal S_t=(G_t,\mathcal L_t,\mathcal M_t,\Pi,\mathcal H_t,\mathcal E_t,\mathcal B_t).
\]
\(G_t=(V_t,E_t)\) is a typed directed multigraph; \(\mathcal L_t\) stores
justification labels; \(\mathcal M_t\) is concise shared memory; \(\Pi\)
contains agents, tools, and policies; \(\mathcal H_t\) is the intervention
ledger; \(\mathcal E_t\) is an immutable evidence/provenance ledger; and
\(\mathcal B_t\) records token, latency, and monetary budgets.

We use \emph{decision graph} for the full typed structure. The 600-instance
mechanism benchmark uses acyclic dependency graphs, hence the term
\emph{decision DAG} in that experiment. A deployed deliberation can contain
cycles through mutual rebuttal, recursive requirements, or version relations.
The propagation engine therefore condenses every strongly connected component
into one supernode and runs over the resulting condensation DAG. This
distinction is essential: the raw workspace is not assumed to be acyclic.

A node is
\(v=\langle id,\tau,x,s,o,c,p,t\rangle\): type, content, status, owner,
confidence, provenance, and timestamp. Types are evidence/fact, assumption,
hard or soft constraint, claim, objection, alternative, risk, question,
intermediate conclusion, and decision. Status is active, weakened, contested,
stale, superseded, rejected, unresolved, or accepted. Edge types include
\(\mathsf{requires}\), \(\mathsf{supports}\), \(\mathsf{derives}\),
\(\mathsf{rebuts}\), \(\mathsf{undercuts}\), \(\mathsf{supersedes}\),
\(\mathsf{answers}\), and \(\mathsf{requiresHuman}\). Evidence nodes require a
source locator, retrieval time, and content hash; model-generated claims never
become evidence merely by repetition.

\begin{table}[t]
\centering
\caption{Typed edges and their revision semantics.}
\label{tab:edges}
\scriptsize
\setlength{\tabcolsep}{3pt}
\begin{tabular}{p{0.15\columnwidth}p{0.68\columnwidth}}
\toprule
\textbf{Relation} & \textbf{Meaning under an upstream change}\\
\midrule
\(\mathsf{requires}\) & Hard premise; loss invalidates that environment.\\
\(\mathsf{supports}\) & Defeasible support; loss weakens or invalidates only
the environments containing it.\\
\(\mathsf{derives}\) & Records the rule/tool and source lineage.\\
\(\mathsf{rebuts}\) & Incompatible conclusion; sets contestation, not staleness.\\
\(\mathsf{undercuts}\) & Attacks an inference or environment's applicability.\\
\(\mathsf{supersedes}\) & Replaces an artifact while retaining history.\\
\(\mathsf{requiresHuman}\) & Blocks synthesis until judgment or explicit deferral.\\
\bottomrule
\end{tabular}
\end{table}

\paragraph{Minimal justification environments.}
For each derived node \(v\), the label
\(\mathcal L_t(v)=\{J_v^1,\ldots,J_v^m\}\) contains subset-minimal alternative
environments. Each \(J\) is a conjunction of supporting literals and rule
identifiers; the label is their disjunction. Let
\[
\mathrm{ok}_t(J)=\bigwedge_{\ell\in J}\mathrm{active}_t(\ell)\ \land\
\neg\mathrm{undercut}_t(J).
\]
Then \(v\) remains supportable iff \(\bigvee_{J\in\mathcal L_t(v)}
\mathrm{ok}_t(J)\). Loss of some but not all environments weakens \(v\);
loss of all makes it stale. A rebuttal makes a claim contested unless an
explicit rule or authoritative evidence invalidates its justification. This
prevents majority agreement from silently deleting a minority objection.

\begin{figure}[t]
\centering
\includegraphics[width=\textwidth]{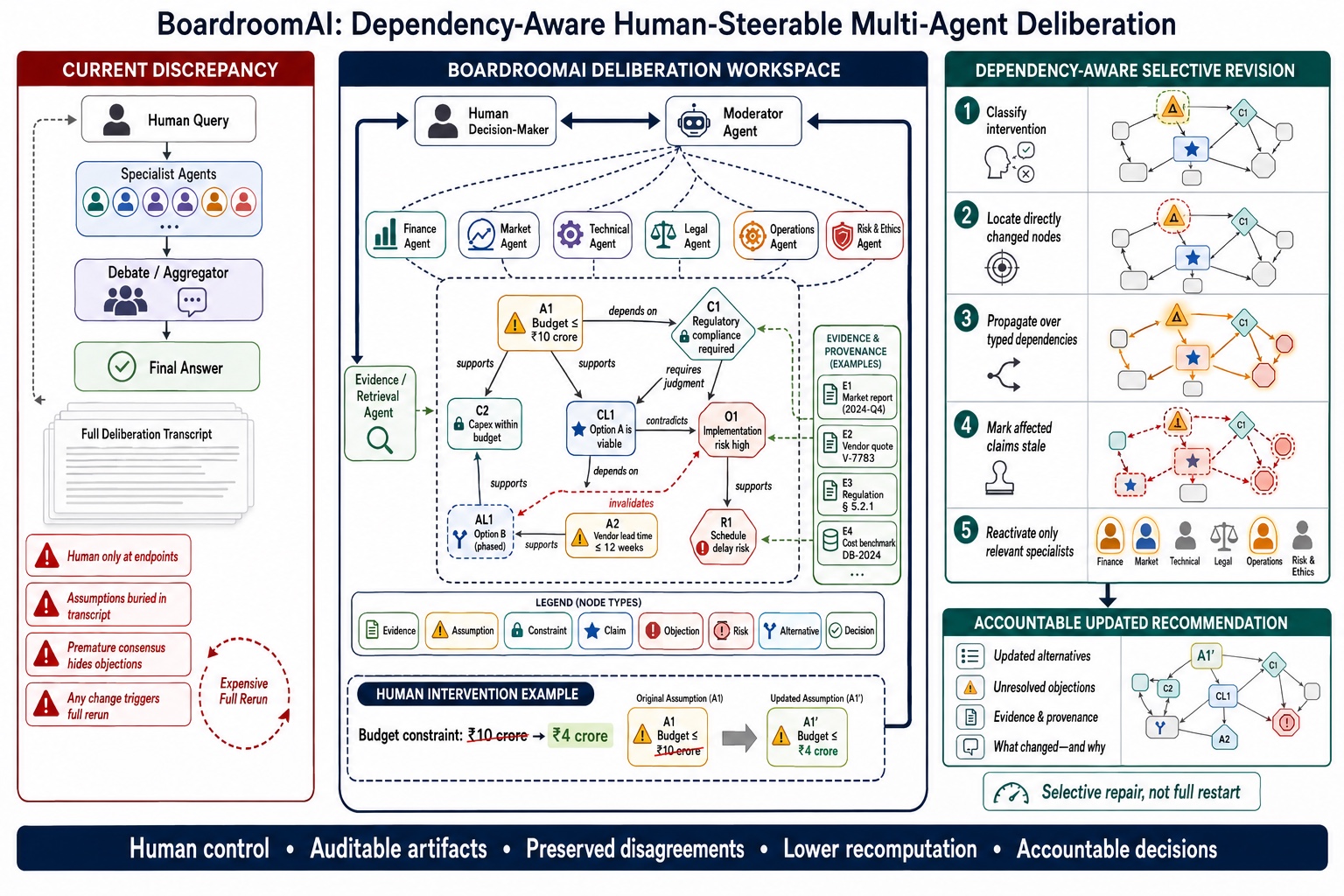}
\caption{BoardroomAI overview. A typed workspace replaces transcript-only
aggregation; confirmed human edits propagate through dependencies and reactivate
only relevant specialists. Exposed objects are decision-support artifacts and
provenance, not private chain-of-thought.}
\label{fig:overview}
\end{figure}

\section{BoardroomAI Framework}
\subsection{Architecture and Deliberation Protocol}
The human can add, retract, replace, challenge, confirm, prioritize, or resolve
an artifact at any turn (\cref{fig:overview}). The moderator compiles
the utterance into candidate typed operations and asks for confirmation when
scope or authority is ambiguous. Specialist agents receive only their role,
the task packet, relevant active subgraph, open objections, evidence excerpts,
and output schema. A retriever returns source-bounded evidence nodes; it cannot
vote. A provenance validator checks locators and quotation entailment. A
skeptic must search for shared assumptions, missing evidence, and counterexamples.

Each specialist emits atomic graph operations rather than an unrestricted
essay: propose/qualify claim, attach evidence, declare assumption, add
justification, object/undercut/rebut, propose alternative, quantify risk, or
ask a question. The deterministic graph service checks node typing, source
availability, cycles, environment minimality, and ownership. The moderator's
turn score combines coverage of open repair obligations, expected information
gain, decision risk, cost, and redundancy. It requests human judgment for
conflicting hard constraints, irreducible value trade-offs, high-sensitivity
near ties, unresolved source conflicts, ambiguous interventions, or unreliable
dependency coverage.

\paragraph{Disagreement and uncertainty.}
Disagreement is relational: two artifacts rebut one another or an objection
undercuts a stated environment. Uncertainty is epistemic: a node's probability
or interval reflects confidence in a checkable proposition. They are not
interchangeable. Each probabilistic claim declares its event, horizon, and
elicitation method; the system aggregates only commensurable forecasts and
retains the individual forecasts. The moderator cannot resolve a normative
conflict by averaging confidence. An objection has a disposition
\(\{\)open, answered, accepted-residual, rejected-with-basis, deferred-by-human\(\}\);
only the human can defer a mandatory value judgment. Minority reports are
generated from open or accepted-residual objections and include the evidence
and condition under which they would change the recommendation.

Stopping requires: no stale artifact on a decision's active justification;
all hard constraints resolved; each objection answered, accepted as residual,
or explicitly deferred; evidence and budget thresholds met; and no mandatory
human question open. The accountable packet contains a recommendation,
alternatives, constraints, sources, assumptions, residual objections and
minority report, calibrated uncertainty, intervention/change log, and
reopening conditions. Otherwise the system abstains or returns an unresolved
decision packet.

\subsection{Human Intervention as Decision-Graph Evolution}
\begin{figure}[t]
\centering
\includegraphics[width=\textwidth]{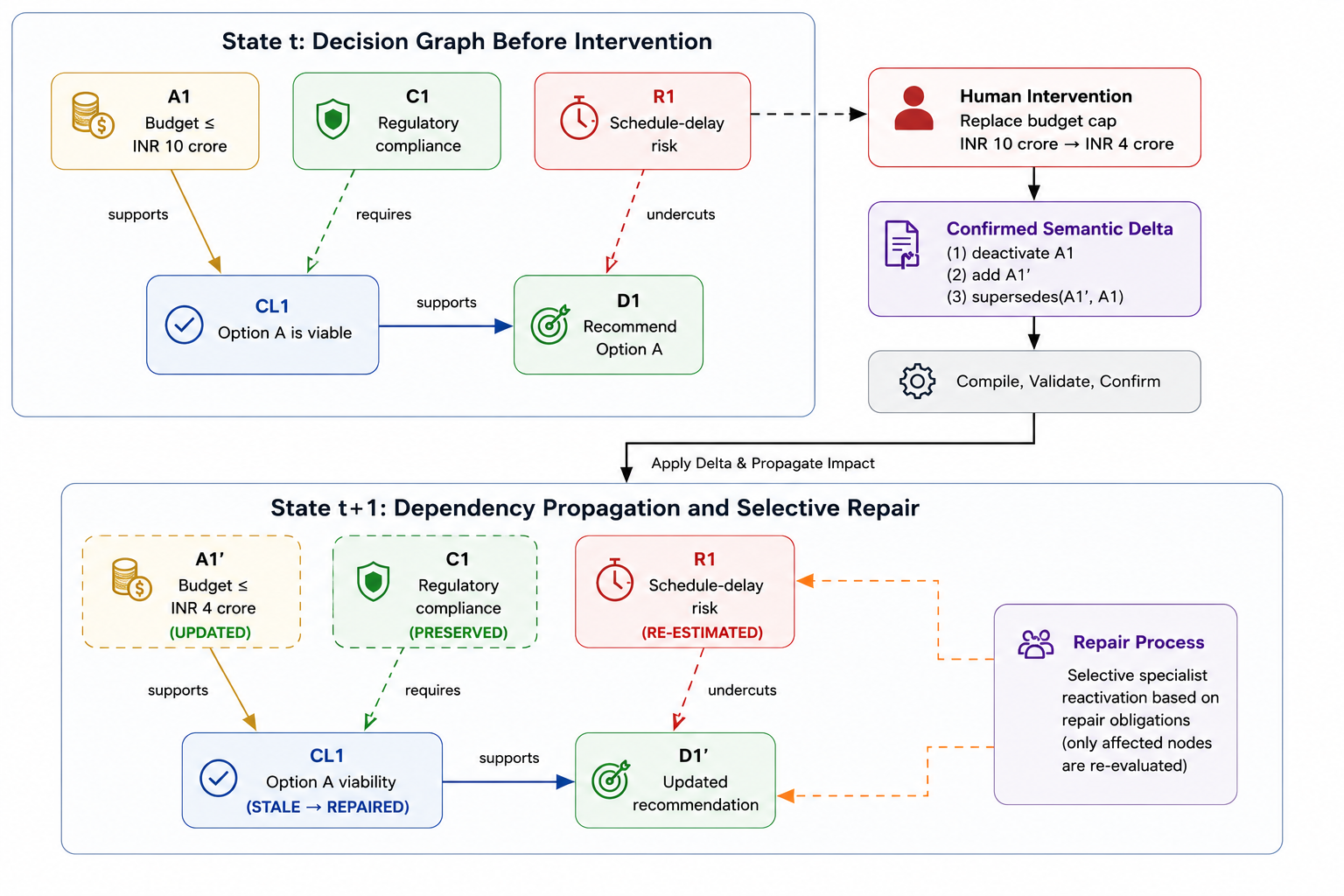}
\caption{A budget revision becomes an explicit semantic delta. Unchanged
regulatory context is preserved; affected viability and schedule artifacts are
repaired. Cyclic workspaces are processed through their SCC condensation DAG.}
\label{fig:evolution}
\end{figure}

An intervention parser returns a confirmed graph delta \(\Delta_h\). Its
components \(D^+\), \(D^-\), \(D^\leftrightarrow\), \(D^?\), and \(\rho\)
encode additions, retractions, replacements, challenges, and
preference/authority updates. A replacement,
for example, does not merely overwrite text: it deactivates the old node, adds
a new version, records a \textit{supersedes} edge, and then reevaluates only
those environments that reference the changed version. \Cref{fig:evolution}
shows this distinction between the direct graph edit and its propagated repair.

Directly changed nodes seed a queue. For each downstream node, the engine recomputes the
semantic signature
\[
q_t(v)=\langle s_t(v),\{J\in\mathcal L_t(v):\mathrm{ok}_t(J)\},
              a_t(v),x_t(v)\rangle,
\]
where \(a_t(v)\) is its active attack set. Only a signature change propagates.
Thus ordinary \(\mathrm{Reachable}^+\) is a safe candidate set but can
over-revise; the impact set \(I_h\) is the least fixed point of actual
signature changes. The repair set \(R_h\subseteq I_h\) contains stale or
contested generated artifacts and content whose source, constraint, or
preference basis changed.

Let \(P_h=V_t\setminus I_h\) denote the \emph{preservation set}. Preservation
does not mean retaining identical wording from a stochastic model; it means
that the validated semantic signature and provenance of an artifact do not
change. Let \(C_h\) be the set of nodes actually recomputed. The system is
selective when \(R_h\subseteq C_h\) and \(|C_h|\ll|V_t|\), and it is safe only
when every mandatory repair obligation is covered or the system falls back to
full redeliberation.

\begin{table}[t]
\centering
\caption{Direct semantics of confirmed human interventions. Downstream status
changes remain justification-dependent.}
\label{tab:interventions}
\scriptsize
\setlength{\tabcolsep}{2.7pt}
\begin{tabular}{p{0.12\textwidth}p{0.34\textwidth}p{0.41\textwidth}}
\toprule
Operation & Direct node update & Edge/semantic effect\\
\midrule
Add & Create a typed, attributed, provenance-linked node. & Add validated relations; reevaluate dependents.\\
Retract & Mark inactive without erasing history. & Exclude from active environments; retain audit edges.\\
Replace & Add a version and supersede the old node. & Add \(\mathsf{supersedes}(v',v)\); preserve replayability.\\
Challenge & Add an objection or counterclaim. & Add \(\mathsf{rebuts}\)/\(\mathsf{undercuts}\); contest, not falsify.\\
Confirm & Record human acceptance. & Update authority only; create no evidence edge.\\
Prioritize & Edit a criterion weight or preference. & Reopen dependent rankings and sensitivities.\\
Override & Add a human decision with stated basis. & Supersede the recommendation; retain residual objections.\\
Resolve & Add an answer or objection disposition. & Add \(\mathsf{answers}\); close only with a valid basis.\\
\bottomrule
\end{tabular}
\end{table}

More formally, let \(\sigma_t(v)\) be the validated local update function for
the node type and let \(\operatorname{pred}(v)\) include support, requirement,
attack, and supersession predecessors. Starting from direct seeds \(D_h\), the
engine computes the least fixed point
\[
I_h=\mu X.\ D_h\cup
\{v:\exists u\in X\cap\operatorname{pred}(v),\
\sigma_{t+1}(v)\ne\sigma_t(v)\}.
\]
The comparison is performed after all alternative environments and active
attacks for \(v\) are recomputed. Within a strongly connected component,
updates iterate to stability; non-convergence or inconsistent hard statuses
forces escalation rather than an arbitrary tie break. Every transition records
the causal predecessor, rule version, old/new signature, and responsible actor.

\begin{algorithm}[t]
\caption{Justification-aware selective redeliberation}
\label{alg:revision}
\begin{algorithmic}[1]
\Require State $\mathcal{S}_t$, confirmed intervention $h$
\Ensure Repaired deliberation state or full redeliberation

\State $\Delta_h \gets
\Call{CompileAndValidate}{h,G_t,\mathcal{E}_t}$

\State $Q \gets \Call{ApplyDirectDelta}{\Delta_h}$
\State $I \gets \emptyset$, $R \gets \emptyset$

\While{$Q \neq \emptyset$}
    \State $u \gets \Call{Pop}{Q}$
    \ForAll{typed dependents $v$ of $u$}
        \State $q' \gets
        \Call{RecomputeSignature}{v,\mathcal{L}_t}$
        \If{$q' \neq q_t(v)$}
            \State update $q_t(v)$ and enqueue $v$ into $Q$
            \State $I \gets I \cup \{v\}$
            \If{$\Call{NeedsRepair}{v,q'}$}
                \State $R \gets R \cup \{v\}$
            \EndIf
        \EndIf
    \EndFor
\EndWhile

\State $O \gets \Call{Obligations}{R,\Delta_h}$

\If{$\Call{GlobalOrUnreliable}{I,G_t}$}
    \State \Return full redeliberation with change log
\EndIf

\State $A' \gets
\Call{BudgetedCover}{O,\Pi,\mathcal{B}_t}$

\State \Return
\Call{DeliberateRepair}{A',O,$\mathcal{S}_t$}
\end{algorithmic}
\end{algorithm}

Obligations include re-estimate, retrieve, reconcile, generate alternative,
adjudicate objection, and resynthesize. Owners, expertise tags, source access,
and estimated cost define a weighted set-cover instance; the router greedily
maximizes uncovered risk-weighted obligations per expected cost, then adds a
skeptic for contested high-impact claims. A full restart is mandatory for
global objective changes, low dependency/extraction confidence, unresolved
cycles, or an affected fraction above a preregistered threshold.

Routing is a budgeted weighted set-cover problem: agent \(a\) has expected
cost \(c_a\), estimated obligation coverage \(K_{ao}\), and each obligation has
risk weight \(w_o\). The online policy greedily maximizes uncovered
risk-weighted coverage per expected cost. Any uncovered mandatory obligation
triggers escalation or fallback, and rejected candidates are logged.

Fallback is based on preregistered diagnostics, not the moderator's rhetorical
confidence: graph-coverage audit score below \(\tau_c\), mean edge-confidence
below \(\tau_e\), changed-node fraction above \(\tau_f\), a global
objective/ontology edit, or an unstable component. Thresholds are tuned only
on development tasks. The evaluation reports a routing safety curve across
deliberately corrupted edges so that cheapness cannot mask incomplete repair.

\paragraph{Conditional preservation.}
Assume complete dependency annotations, deterministic pure update functions,
and propagation over an acyclic graph or its strongly connected component
condensation. Then every node outside the least-fixed-point impact set preserves
its semantic signature. The proof is induction over topological order: an
unchanged node has neither a direct delta nor a changed predecessor signature,
so its deterministic update is unchanged. This is a conditional systems
property, not a claim about fallible LLM extraction. The structural pass is
\(O(|V|+|E|+\sum_v|\mathcal L(v)|)\); LLM cost remains empirical.

\subsection{Decision-Sufficient Repair Packets}
Correct impact routing is necessary but not sufficient. The synthesizer must
also receive enough unchanged boundary context to prove feasibility and
reconstruct the declared decision rubric. We define the
\emph{decision-sufficient closure} of a repair set as
\[
\operatorname{DSC}(R_h)=R_h\cup B_h^{\mathrm{hard}}\cup B_h^{\mathrm{rubric}}
\cup B_h^{\mathrm{alts}}\cup B_h^{\mathrm{risk}}\cup B_h^{\mathrm{source}},
\]
where the five boundary sets contain, respectively, every active hard
constraint relevant to a surviving alternative, every criterion and
human-approved weight used by the synthesizer, the roots of surviving
alternative justifications, unresolved objections and material risks, and the
minimal evidence spans needed to verify included claims. The packet contains
semantic summaries plus identifiers and provenance, not an unrestricted replay
of the meeting.

This closure is a design correction motivated by the reported pilot. The
prototype selective condition supplied affected subgraphs and hand-selected
boundary premises, yet abstained in six of twelve cases because the
synthesizer could not establish a complete constraint/rubric basis. The
reported numbers therefore evaluate the earlier packet construction, not the
full \(\operatorname{DSC}\) rule. A future replication must compare the original
packet, the closure above, and full structured repair under matched budgets.

\subsection{Decision Synthesis and Implementation Blueprint}
Synthesis is constraint-first, not a vote over prose. For alternative \(d\),
the graph service evaluates executable hard predicates
\(\mathbf g(d)\), then computes a transparent soft-utility interval
\[
U(d)=\sum_k \omega_k u_k(d)-\sum_r \eta_r\,p_r(d)\ell_r(d),
\]
where weights \(\omega_k\) are human-approved, \(u_k\) are normalized criterion
scores, and each risk has probability \(p_r\), loss \(\ell_r\), and severity
weight \(\eta_r\). Missing quantities remain intervals or unresolved nodes;
they are not imputed by the judge. The synthesizer ranks only hard-feasible
alternatives, reports sensitivity to weights and uncertain inputs, and
abstains if no alternative is feasible. Agent votes are retained as provenance
but have no direct coefficient in \(U\). This separation lets the benchmark
distinguish a correct calculation from an unsupported preference.

The graph service accepts schema-valid, versioned transactions that name every
premise ID. It verifies identifiers, permissions, immutable evidence spans, and
environment minimality; accepted writes store replayable before/after hashes.
Meeting prose is only a view of this authoritative graph. Specialists declare
premises and edge types, a separate critic tests counterfactual dependence, and
rule validators enforce types; disagreement lowers edge confidence and broadens
routing. The moderator sees graph summaries and obligations, not hidden
reasoning, and every action must cite the obligation or stop condition it
addresses. Evidence prompt injection is treated as quoted data and credentials
remain outside specialist context. These are implementation controls, not
outcomes tested by the present experiments.

\section{Experiments}
We report deterministic propagation on complete generated decision DAGs, a missing-edge stress test with fallback disabled, and a 12-case exploratory agent
pilot. The pilot uses confirmed field-level updates and a non-LLM router; it
does not test unrestricted parsing, human outcomes, live retrieval, or real
organizations. The proposed decision-sufficient closure is not yet evaluated.

\subsection{Mechanism-Level Synthetic Validation}
\paragraph{Scope and setup.}
We implemented the deterministic propagation core in Python and tested it on
600 generated decision DAG interventions (150 each with 64, 128, 256, and 512
nodes). Each derived node had one to three subset-minimal alternative
environments. One or two active source nodes were invalidated; 175 generated
attempts with trivial or near-global impact were rejected before obtaining the
fixed 600-case suite. Seed 20270721, per-case records, the generator, and the
analysis script are included. An independent exhaustive evaluator recomputed
every node and defined gold semantic impact as any change in active environment
IDs. This experiment used no LLM, evidence packet, decision rubric, or human;
it isolates routing semantics and cannot answer RQ1, RQ3, RQ4, or RQ5.

Conditions were direct-node updating; an ablation retaining only one
environment per claim; descendant reachability; exhaustive full restart;
BoardroomAI signature propagation; and an oracle that inspects exactly the gold
impact set. \emph{Recompute} is the fraction of nodes inspected. \emph{Reuse}
is the fraction of gold-unaffected nodes not recomputed, not a claim that a
stochastic LLM would reproduce identical prose. \cref{tab:synthetic}
reports case means; intervals below use 1,000 case-bootstrap draws.

\begin{table}[t]
\centering
\caption{Mechanism-level results on 600 generated interventions. BoardroomAI
uses complete dependency annotations. Values are percentages.}
\label{tab:synthetic}
\setlength{\tabcolsep}{2.5pt}
\begin{tabular}{lrrrrrr}
\toprule
\textbf{Method} & \textbf{P} & \textbf{R} & \textbf{F1} & \textbf{Exact} & \textbf{Inspect} & \textbf{Reuse}\\
\midrule
Direct only & 100.00 & 23.32 & 36.65 & 0.00 & 0.95 & 100.00\\
Single env. & 34.04 & 78.58 & 38.60 & 1.33 & 19.02 & 84.19\\
Reachability & 10.47 & 100.00 & 17.08 & 0.33 & 59.78 & 42.66\\
Full restart & 100.00 & 100.00 & 100.00 & 100.00 & 100.00 & 0.00\\
\textbf{BoardroomAI} & \textbf{100.00} & \textbf{100.00} & \textbf{100.00} &
\textbf{100.00} & \textbf{14.59} & \textbf{89.96}\\
Oracle & 100.00 & 100.00 & 100.00 & 100.00 & 5.75 & 100.00\\
\bottomrule
\end{tabular}
\end{table}

With complete annotations, selective propagation agreed with exhaustive
recomputation in all 600 cases. It inspected 14.59\% of nodes (95\% bootstrap
CI 13.47--15.85) and reused 89.96\% of unaffected nodes (89.11--90.80).
Reachability obtained full recall by inspecting 59.78\% of nodes, but its mean
precision was 10.47\% (9.49--11.50). Direct updating and a single-environment
label were cheaper but missed downstream or alternative-environment changes.
The oracle gap (14.59\% versus 5.75\%) is boundary-inspection overhead: the
selective algorithm must inspect candidates to prove they are unchanged.
Median Python structural speed-up over full recomputation rose from 2.11$\times$
at 64 nodes to 22.08$\times$ at 512 nodes; these timings reflect local synthetic
edits on one CPU, not LLM latency or token savings.

\paragraph{Missing-dependency stress test.}
We randomly deleted routing edges while retaining the complete graph for gold
evaluation and deliberately disabled fallback (240 cases per rate). As
\cref{tab:corruption} shows, only 10\% deletion reduced recall to 90.55\%
(88.48--92.52) and exact-set recovery to 60.42\%. Precision remains 100\%
because deletion hides paths but cannot create false ones in this monotone
simulation; wrong-edge corruption remains untested. Thus the perfect complete-
graph result validates an implementation conditional, while the stress test
exposes dependency extraction and fallback detection as unresolved risks.

\begin{table}[t]
\centering
\caption{Observed selective routing without fallback under random edge
deletion (240 cases per rate; percentages).}
\label{tab:corruption}
\setlength{\tabcolsep}{4.1pt}
\begin{tabular}{rrrrr}
\toprule
\textbf{Deleted} & \textbf{Precision} & \textbf{Recall} & \textbf{F1} & \textbf{Exact set}\\
\midrule
0\% & 100.00 & 100.00 & 100.00 & 100.00\\
5\% & 100.00 & 96.11 & 97.68 & 81.25\\
10\% & 100.00 & 90.55 & 94.19 & 60.42\\
20\% & 100.00 & 85.09 & 90.86 & 47.08\\
30\% & 100.00 & 75.13 & 83.74 & 23.75\\
\bottomrule
\end{tabular}
\end{table}

\subsection{Exploratory Validation}

The primary contribution of BoardroomAI is a methodology for dependency-aware,
human-steerable multi-agent deliberation rather than a new language model or
benchmark. Consequently, our objective is not to establish state-of-the-art
decision quality, but to verify that the complete intervention--routing--repair
pipeline can be instantiated using contemporary LLM agents and executed
end-to-end under controlled conditions. To this end, we conduct an exploratory
validation using a synthetic organizational benchmark with hidden interventions.
The study evaluates whether selective repair preserves unaffected reasoning,
correctly propagates intervention effects, and reconstructs updated decisions
without unnecessary re-deliberation. Equally importantly, it serves as a
feasibility study for identifying practical failure modes and guiding the design
of future large-scale evaluations.

\paragraph{Exploratory Codex validation.}
The validation consists of an end-to-end stress test over a controlled
12-task synthetic benchmark. Each organizational case was constructed before
generation and contains four candidate decisions, supporting evidence,
executable hard constraints, weighted decision criteria, and one hidden
intervention that alters both the feasible solution space and the optimal
decision. Ground-truth dependency annotations identify directly affected,
transitively affected, revision-required, and unaffected artifacts, allowing
repair accuracy to be evaluated automatically.

Generation agents received only phase-specific public information, while a
deterministic validator verified every ground-truth label. For the structured
conditions, a canonical decision graph was generated mechanically from the
public artifacts. Following revelation of the hidden intervention, a
non-LLM controller propagated changes through typed dependency edges to
identify the minimal repair region. This evaluates selective repair assuming an
available dependency graph rather than the separate problem of dependency
extraction.

Four execution settings were compared. C1 employs a single iterative agent.
C2 uses multiple specialist agents followed by complete re-deliberation after
every intervention. C3 performs structured full re-deliberation using the
canonical decision graph, while C4 activates only specialists whose artifacts
lie within the routed repair region. The exploratory implementation was
executed using the product-facing \texttt{gpt-5.6-terra} interface with medium
reasoning. Since immutable model snapshots, decoding controls, seeds, token
usage, hardware configuration, and latency statistics were unavailable, this
study is presented as an exploratory feasibility validation rather than a fully
reproducible benchmark.

\begin{table}[t]
\centering
\caption{Exploratory agent results. Percentages are task-run means. Utility was
100\% conditional on a non-abstaining valid choice in every condition. Unequal
repetitions preclude a superiority claim.}
\label{tab:pilot}
\setlength{\tabcolsep}{2.0pt}
\begin{tabular}{lrrrrrrr}
\toprule
\textbf{Condition} & \textbf{\(n\)} & \textbf{Valid} & \textbf{Impact F1} & \textbf{Rev. F1} & \textbf{Preserve} & \textbf{Inspect} & \textbf{Exact}\\
\midrule
C1 iterative & 36 & 100.00 & 61.44 & 65.28 & 45.24 & 25.88 & 0.00\\
C2 full panel & 36 & 100.00 & 43.77 & 38.44 & 0.00 & 42.51 & 0.00\\
C3 struct. full & 12 & 100.00 & 100.00 & 100.00 & 0.00 & 100.00 & 100.00\\
C4 selective & 12 & 50.00 & 100.00 & 100.00 & 100.00 & 62.11 & 50.00\\
\bottomrule
\end{tabular}
\end{table}

\paragraph{Quantitative results.}
C1, C2, and C3 produced a valid updated decision for every evaluated task,
reflecting the transparency of the synthetic benchmark rather than real-world
decision quality. C1 correctly identified only 44.44\% of affected artifacts,
whereas C2 preserved none of the gold-unaffected reasoning because every
specialist was restarted. C3 achieved perfect routing accuracy but recomputed
every canonical artifact, including 37.89\% that should have remained
unchanged. In contrast, C4 preserved every unaffected artifact and reduced
specialist activations by 11.11\%, recomputing only 62.11\% of canonical
artifacts. However, it abstained on six of the twelve tasks, producing a valid
updated recommendation only when the routed repair packet contained sufficient
information to reconstruct the final decision.

\paragraph{Qualitative analysis.}
The most important observation is that dependency-correct routing alone is
insufficient for successful selective repair. Although C4 achieved perfect
routing and preservation on all evaluated tasks, it abstained whenever the
available repair context was insufficient to establish a complete
constraint--criterion basis for the synthesizer. This suggests that future
selective-repair systems should optimize not only routing accuracy but also
\emph{decision sufficiency}, ensuring that repaired artifact subsets contain
enough information to reconstruct globally consistent recommendations.

\paragraph{Scoring and limitations.}
Programs compared predicted decisions and artifact sets against frozen private
labels. Decision quality, preservation, unnecessary recomputation, and
routing precision/recall/F1 were computed automatically using exact canonical
identifiers. Confidence intervals were estimated using 5,000 task-cluster
bootstrap samples. Because structured conditions completed only a single paired
execution and backend reproducibility controls were unavailable, no statistical
superiority claims are made. Risk descriptions, factual entailment, confidence
calibration, execution cost, latency, and token consumption were therefore
excluded from quantitative analysis.

\paragraph{Future evaluation.}
A comprehensive evaluation should include independently adjudicated decision
cases, multiple model families, controlled decoding configurations, matched
token budgets, oracle and full-repair baselines, and ablation studies of typed
dependencies, provenance, escalation, alternative justification environments,
and routing policies. Robustness experiments should additionally evaluate
incorrect dependency graphs, ambiguous interventions, conflicting constraints,
fabricated evidence, and fallback strategies. The principal safety objective is
to demonstrate that selective repair preserves decision quality while reducing
unnecessary re-deliberation.

\section{Design Implications for Human--AI Co-Creation}
Human steerability should update authoritative state rather than append another
message. Preservation is also a co-creative capability because it protects
valid alternatives, dissent, and provenance from unnecessary stochastic
rewriting. Finally, routing minimality and decision sufficiency are different:
a small impact set may still omit the hard constraints or rubric context needed
for synthesis. Evaluation must therefore report validity, preservation,
coverage, completion, and cost together.

\paragraph{Planned human evaluation.}
A counterbalanced within-subject study will compare iterative single-agent,
structured full, and selective repair. Its primary endpoint is change-basis
comprehension: what changed, what was preserved, and why, with task score,
calibrated reliance, time, control, situation awareness, and NASA-TLX as
secondary outcomes \cite{hart1988tlx,schemmer2023reliance}. No participant result
is claimed.

\section{Reproducibility}\label{sec:repro}
The specified supplement contains the structural runner, fixed seed, 600 graph
records, corruption records, 12-case generator and validator, frozen prompts,
outputs, canonical labels, exclusion ledger, scorer, bootstrap summaries, and
hashes. Unavailable backend fields remain null; the interrupted repetition is
listed but excluded before synthesis.

\section{Conclusion}
BoardroomAI models human--AI deliberation as evolution of an external decision
graph: typed human edits propagate through alternative justifications, trigger
obligation-based specialist repair, and preserve unaffected objections and
provenance. On 600 complete generated decision DAGs, propagation exactly matched
exhaustive impact sets while inspecting 14.59\% of nodes; missing edges sharply
degraded recovery. In the 12-case pilot, selective repair preserved all
unaffected canonical nodes but abstained on half the cases. This negative result
motivates decision-sufficient closure: correct routing must also supply the
constraints, rubric, alternatives, risks, and evidence needed for synthesis.
The contribution is therefore a falsifiable mechanism, not evidence of general
organizational superiority.

\section*{Acknowledgements}

We thank the anonymous reviewers for their valuable comments and constructive
feedback. We also thank colleagues and mentors for insightful discussions on
multi-agent systems, computational argumentation, and human--AI collaboration
that helped shape this work.

%
%
\bibliographystyle{splncs04}
\bibliography{main}
\end{document}